\documentclass[hyperref]{article}
\usepackage{lineno,hyperref}
\modulolinenumbers[5]

\usepackage[a4paper,left=18mm,right=18mm,top=15mm,bottom=15mm]{geometry}
\usepackage{amsmath,amssymb,amsfonts}
\usepackage{graphicx}
\usepackage{textcomp}
\usepackage{xcolor}
\usepackage{diagbox}
\usepackage{subfigure}
\usepackage{booktabs}
\usepackage{multirow}
\usepackage{makecell}
\usepackage{color}
\usepackage{float}
\usepackage{url}
\usepackage{booktabs}
\usepackage{cleveref}
\usepackage{mathrsfs}

\usepackage{bm}
\usepackage{textcomp} 
\usepackage{bbding} 
\usepackage{stfloats} 
\usepackage{diagbox} 

\usepackage[linesnumbered,ruled,boxed]{algorithm2e}
\usepackage{algpseudocode}

\usepackage{authblk}
\usepackage{amsthm}
\theoremstyle{definition}
\newtheorem{definition}{Definition}

\def\BibTeX{{\rm B\kern-.05em{\sc i\kern-.025em b}\kern-.08em
    T\kern-.1667em\lower.7ex\hbox{E}\kern-.125emX}}



\begin{document}


\title{\textbf{DP$^2$-NILM: A Distributed and Privacy-preserving Framework for Non-intrusive Load Monitoring}}
%

%
\author[1]{Shuang Dai\thanks{sd19628@essex.ac.uk}}
\author[1]{Fanlin Meng\thanks{fanlin.meng@essex.ac.uk}}
\affil[1]{Department of Mathematical Sciences, University of Essex, Colchester, UK}
\author[2]{Qian Wang\thanks{qian.wang173@hotmail.com}}
\affil[2]{Department of Computer Science, Durham University, Durham, UK}
\author[3]{Xizhong Chen\thanks{xizhong.chen@sheffield.ac.uk}}
\affil[3]{Department of Chemical and Biological Engineering, University of Sheffield, UK}
\date{June 2022}

\maketitle



\begin{abstract}
  Non-intrusive load monitoring (NILM), which usually utilizes machine learning methods and is effective in disaggregating smart meter readings from the household-level into appliance-level consumption, can help analyze electricity consumption behaviours of users and enable practical smart energy and smart grid applications. Recent studies have proposed many novel NILM frameworks based on federated deep learning (FL). However, there lacks comprehensive research exploring the utility optimization schemes and the privacy-preserving schemes in different FL-based NILM application scenarios. In this paper, we make the first attempt to conduct FL-based NILM focusing on both the utility optimization and the privacy-preserving by developing a distributed and privacy-preserving NILM (DP$^2$-NILM) framework and carrying out comparative experiments on practical NILM scenarios based on real-world smart meter datasets. Specifically, two alternative federated learning strategies are examined in the utility optimization schemes, i.e., the FedAvg and the FedProx. Moreover, different levels of privacy guarantees, i.e., the local differential privacy federated learning and the global differential privacy federated learning are provided in the DP$^2$-NILM. Extensive comparison experiments are conducted on three real-world datasets to evaluate the proposed framework.\\
  \textbf{\textit{keywords:}} deep neural network, differential privacy, federated learning, non-intrusive load monitoring, privacy-preserving
\end{abstract}

\section{Introduction}
\subsection{Background and Motivations}
Modern urbanization, lifestyles, and technological advancements have increased the energy demand. Energy supply generates greenhouse gas emissions that accelerate climate change, which presents a significant threat to the security and prosperity of the global community \cite{carbon_plan}. In the UK, legal obligations regarding climate change have been enacted, putting increased strain on the traditional centralized power grid \cite{connor2014policy}. The smart grid has been brought up using information systems to create a more reliable and intelligent power grid network \cite{lamnatou2021smart}, which has the potential to contribute to the decarbonization of the energy system and is a leading candidate for renewable energy sources \cite{connor2014policy}. As a key part of a smart grid, smart meters allow non-intrusive appliance load monitoring \cite{hart1992nonintrusive} to help smart meter clients reduce energy consumption by scheduling appliance usage hours and monitoring abnormal electricity usage patterns.

Non-intrusive load monitoring (NILM) \cite{hart1992nonintrusive} is a growing trend in utilizing machine learning methods to monitor events (ON/OFF) or energy consumption of individual appliances using the aggregated smart meter reading of the whole building \cite{gopinath2020energy}. NILM provides real-time feedback regarding the energy consumption of smart meter clients, and research findings indicate that the appliance-level feedback may be able to save up to 12\% of annual energy consumption \cite{armel2013disaggregation}. 

Deep learning-based models have presented new opportunities for the electrical utility industry \cite{MISHRA2020104000}, and are the most representative structures applied to NILM \cite{kelly, yuan39, gopinath2020feature, kukunuri2020edgenilm}, which have been proved to be more effective than other traditional models. However, most deep learning-based NILM models are centralized \cite{meidan2018n}, which may not be feasible in the era of big data due to data privacy concerns and excessive communication overhead from numerous smart meter devices.
To address these challenges, researchers have used federated learning (FL) \cite{dianxin1-Communication}, an emerging paradigm for training models that can be tailored to individuals without relying on centralized data \cite{kairouz2021advances}. Furthermore, FL-based NILM models benefit from the collaboration of multiple data sources and privacy guarantees, which are more efficient than models trained solely on individual households. Even though the FL paradigm has obvious advantages for NILM, there are still challenges in real-world applications:
\begin{itemize}
    \item When the parameters of the model are exchanged between the central server and local clients, FL has been identified as being vulnerable to privacy invasions \cite{kairouz2021advances, bagdasaryan2020backdoor}.
    \item In practice, different households use energy in different ways, which makes current FL-based NILM models ineffective for dealing with the heterogeneity of smart meter clients \cite{d2019transfer}.
\end{itemize}

\subsection{Contributions}
Current FL-based NILM models either rely on the naive federated learning framework without any additional privacy guarantees, or only attempt to address the above challenges from a single viewpoint, i.e., adopt a single privacy-preserving scheme for FL-based NILM or optimize the FL algorithm solely to address the heterogeneity challenge. However, in practical real-world NILM scenarios, different smart meter clients may have different requirements, making it challenging for FL-based models to meet all of these demands.
There lack comprehensive experimental studies on systematically understanding different enhancement schemes for FL-based NILM concerning the above challenges in practical application scenarios. To this end, this paper presents a DP$^2$-NILM framework to address the aforementioned challenges by exploring both utility optimization and the privacy preservation schemes for a variety of FL-based NILM scenarios. The main contributions of the paper are summarized as follows. 
\begin{itemize}
	\item We propose DP$^2$-NILM, a distributed and privacy-preserving framework by deploying federated learning with two enhancing schemes, i.e., the utility optimization and the privacy-preserving in real-world NILM scenarios. The DP$^2$-NILM is expected to provide a systematic understanding of different enhancement schemes for the challenges in real-world FL-based NILM applications to satisfy the diverse requirements of smart meter clients.
	
	\item To deal with client heterogeneity, the DP$^2$-NILM framework examines two utility optimization schemes derived from the FL paradigm, the Federated averaging (FedAvg) \cite{dianxin1-Communication} and FedProx \cite{li2020federated}. By exploring how the utility optimization schemes impact the accuracy of the FL-based NILM models, the proposed DP$^2$-NILM can achieve satisfactory performance under both the homogeneous and heterogeneous data environments.
	
	\item The DP$^2$-NILM framework preserves data privacy by incorporating differential privacy (DP) at different levels, i.e., the central DP level and local DP level, and aims to find a better trade-off between utility and privacy for various privacy requirements.
	
	\item We examine the correlations between key parameters and the inference accuracy of different enhancement schemes. It has been demonstrated that the proposed DP$^2$-NILM can be scalable and that it offers enlightening insights into different smart meter client requirements based on the extensive evaluation of three real-world datasets.
\end{itemize}

The remainder of this paper is structured as follows. Section \ref{RW} reviews literature related to the proposed DP$^2$-NILM framework. Section \ref{BP} provides background knowledge and briefs the preliminaries used in DP$^2$-NILM. Section \ref{DP2NILMFramework} overviews the three-tier workflow of the proposed DP$^2$-NILM. The utility optimizations schemes and the privacy-preserving schemes of $DP^2$-NILM are detailed in Section \ref{UoDP2NILM} and Section \ref{PpDP2NILM}, respectively. The performance evaluations on real-world datasets are conducted in Section \ref{PE}. The conclusion and possible future extensions are given in Section \ref{C}.

\section{Related Work} \label{RW}

\subsection{NILM and its privacy-preserving methods}
NILM was firstly proposed by Hart \cite{hart1989residential}, which utilized heuristic methods based on Combinatorial Optimization (CO) to perform load disaggregation. After this, diverse models have been proposed to improve inference accuracy, which can be mainly divided into supervised learning and unsupervised learning approaches. For the former, deep neural network (DNN) based models \cite{zhang2018sequence, shin2019subtask, bejarano2019deep}, which provide new opportunities for the electrical utility industry \cite{MISHRA2020104000}, are the most representative structures applied to NILM. For unsupervised learning, the factorial Hidden Markov Model (FHMM) \cite{yuan26, xia2021non, salem2021unsupervised} and clustering analysis \cite{yuan49} are commonly used for NILM. For instance, \cite{salem2021unsupervised} develops a new infinite factorial hidden Markov model for NILM constrained on contextual features, which utilizes the usage information on the appliance-level to improve the disaggregation accuracy. \cite{kelly} compared unsupervised learning (CO, FHMM) with supervised learning (DNN) for NILM, where the DNN-based model achieved the best performance.

DNN-based NILM models depend heavily on diverse and abundant training data, yet real-world datasets are often isolated. Moreover, due to the communication bandwidth limitation and the data privacy legislation, it is difficult to integrate smart meter readings from different buildings into a centralized database. On the other hand, the emergence of federated learning \cite{li2019federated} not only provides privacy guarantees for smart meter data but solves the challenge of data isolation, which therefore brings considerable benefits to DNN-based NILM models. Despite its promising future, applying FL to NILM has only received attention very recently  \cite{dai2021federatednilm, wang2022fed, zhang2022fednilm, potter2021towards}. For instance, \cite{dai2021federatednilm} proposed a FederatedNILM framework to enable NILM task in the FL paradigm at the residential level. \cite{wang2022fed} utilized the FL paradigm to improve the model performance for NILM in both residential and industrial scenarios. \cite{zhang2022fednilm} proposed a FedNILM framework utilizing model compression to reduce the computation overhead while retaining satisfying performance for NILM. \cite{potter2021towards} adopted DP into the FL for NILM to provide stronger privacy protection, and a membership attack was included to evaluate the privacy guarantee level of the framework. 

\subsection{Enhancing mechanisms of federated learning}
There are two main streams of approaches for enhancing the FL framework, i.e., the utility optimization schemes and the privacy-preserving schemes. 

In recent years, many advanced utility optimization schemes have been proposed \cite{dianxin1-Communication, nguyen2020fast, khan2020federated, kang2019incentive, multi, network}. For example, \cite{multi} used FL for multi-task network anomaly detection, which improved the training efficiency compared with multiple single-task models. Later, \cite{network} combined FL with the deep neural network to solve the similar problem. Moreover, transfer learning was adopted in this paper to reconstruct the model for improving the anomaly detection performance. Among them, the most commonly used mechanism is the federated averaging \cite{dianxin1-Communication}, which averages the updated gradients from the client models to optimize the global model. However, the FedAvg has been demonstrated to diverge empirically in scenarios where the data is non-independent and identically distributed (non-IID) across clients \cite{dianxin1-Communication}. FedProx \cite{li2020federated}, which uses proximal terms to stabilize model updating, was then proposed as a solution to heterogeneity in federated networks.

Although many utility optimization mechanisms have been proposed, FL offers limited privacy guarantees. Prior study has proposed differential private FL (DPFL) to provide clients with stronger privacy guarantees, which has been used as the basis for many privacy-preserving FL-based schemes \cite{cao2020ifed, hudson2021framework, potter2021towards, wang2020privacy}. Privacy in FL can be divided into global differential privacy FL (GDPFL) \cite{wei2020federated} and local differential privacy FL (LDPFL) \cite{cao2020ifed}
based on different noise adding mechanisms. In GDPFL, the trusted server applies the noise during the parameter aggregation, whereas in LDPFL, each participant adds noise to the model parameters before uploading them to the server.

Most existing studies only provided privacy guarantees for FL-based NILM at a fixed level, and little attention was paid to exploring privacy-preserving schemes for FL on different scales, e.g., at both the global and local levels. Moreover, while the data heterogeneity in FL-based NILM scenarios is a practical and important characteristic due to different users have varied lifestyles and, accordingly, different electricity usage patterns, there is no existing research having tackled this challenge from the utility optimization perspective. Therefore, in this paper, we make the first attempt to explore FL-based NILM focusing on the utility optimization schemes and the privacy-preserving schemes by developing the DP$^2$-NILM framework and conducting extensive and comparative experiments on practical NILM scenarios based on real-world smart meter datasets.

\section{Preliminaries} \label{BP}

We introduce several essential concepts related to the proposed DP$^2$-NILM framework in this section.

\subsection{Non-intrusive load monitoring}
Given the aggregated load $L_t$ at time $t$:
\begin{equation}\label{eq1}
L_t=\sum_{i=1}^Il_t^{i}+\gamma_t,
\end{equation} 
the goal of non-intrusive load monitoring (NILM) is to recover the status of $I$ target electrical appliances. $l_t^i$ and $\gamma_t$ denote the load consumption for the $i$-th appliance and the residual/unmonitored load respectively at time $t$. NILM can be formulated as either a classification task or a regression task depending on the status variables of individual electrical appliances we aim to recover.

For the regression task, the NILM model aims to find the approximation, denotes as $F$, of the true relationship between the aggregated household-level consumption ($L_t$) and the appliance-level consumption  
\begin{equation}\label{eq2}
\pmb{L}=[\hat{l}_t^{1},\hat{l}_t^{2},\ldots, \hat{l}_t^{i}, \ldots,\hat{l}_t^{I}] = F(L_t),
\end{equation} where $\pmb{L}$ is the predicted load consumption sequence of $I$ target electrical appliances at time $t$.

For the classification task, thresholds need to be set for the NILM model to determine the states (e.g., ON/OFF) of each target appliance. A commonly used threshold method is the activation-time thresholding, which could avoid the negative effect of the abnormal spikes during the OFF state to better improve the inference accuracy \cite{kelly}. For the sake of simplicity, we assume that there are two typical states (ON/OFF) for the target appliances, and the state $s_t^i$ for $i$-th appliance at time $t$ is related to its threshold $\lambda^i$
\begin{equation}
s_t^i=
\left\{\begin{matrix}
1,& {l_t^i \geq \lambda^i}\\
0,& {l_t^i < \lambda^i},
\end{matrix} \right.
\end{equation}
where 0 represents the OFF state, and 1 denotes the ON state. Therefore the classification task for NILM can be defined as
\begin{equation}\label{eq3}
\pmb{S}=[\hat{s}_t^{1},\hat{s}_t^{2},\ldots, \hat{s}_t^{i}, \ldots,\hat{s}_t^{I}] = F_s(L_t),
\end{equation} 
where $\hat{s}_t^{i}$ is a binary variable indicating the predicted ON/OFF state of $i$-th electrical appliance at time $t$.

\subsection{Federated deep learning}

When data owners intend to combine their local data to train a common utility model, the traditional centralized approach is to pool their own private data at a central server, during which the data uploading and integration process are often restricted by data privacy legislation. To address this challenge, FL was brought up \cite{li2019federated}, which only requires the exchange of updated model parameters rather than the raw data between clients and the central server, and therefore is deemed to be the state-of-the-art approach for distributed data privacy protection. 

\begin{figure}[!htbp]
	\centering
	\includegraphics[width=0.5\textwidth]{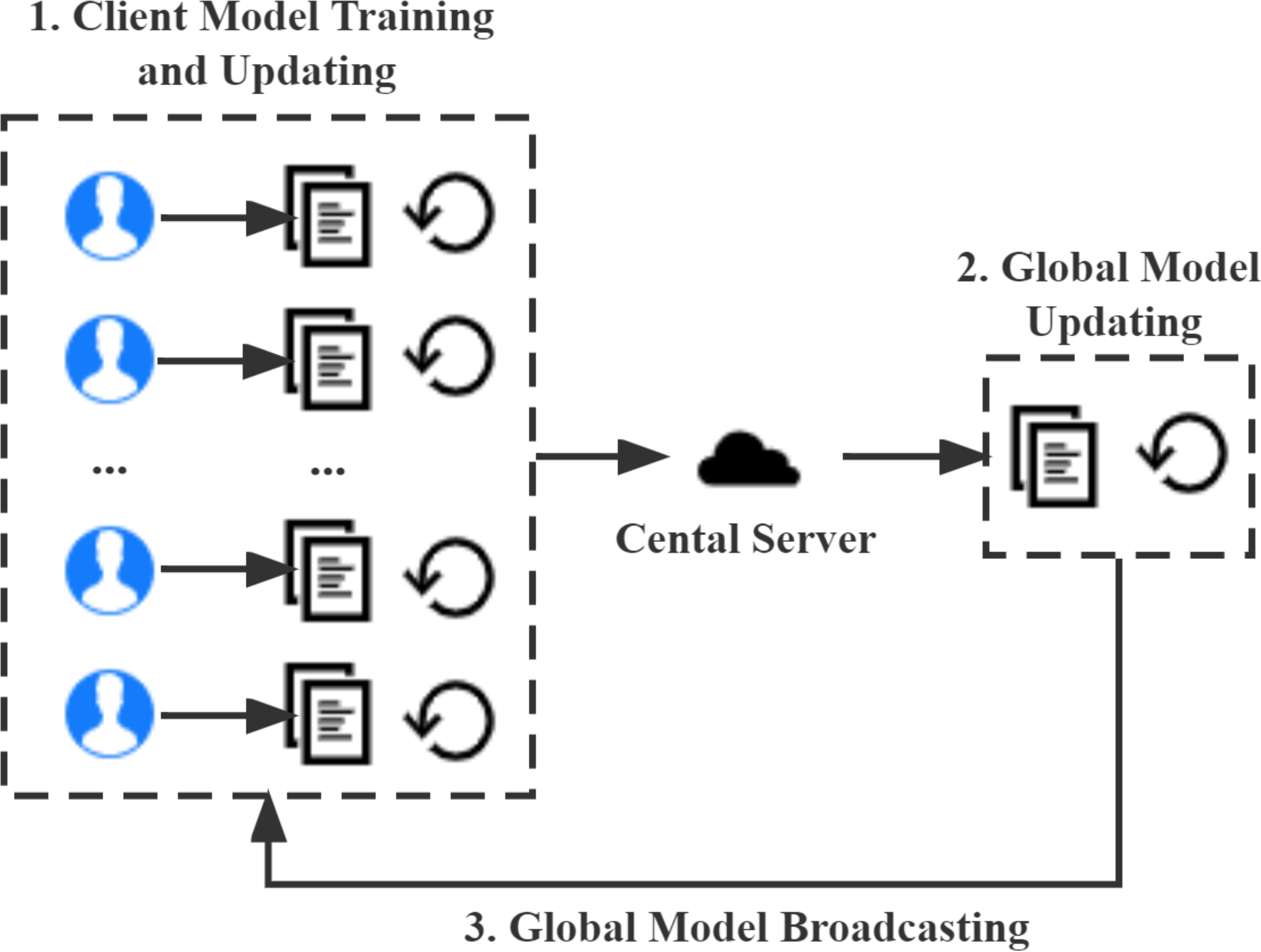}
	\caption{Training process of the federated deep learning framework.}
	\label{DP2NILMFL}
\end{figure}

FL is a machine learning strategy aimed at training a high-quality global model while the raw private datasets are distributed locally in each client without the need for transferring them to a central server. The training process of the federated deep learning framework is shown in Figure \ref{DP2NILMFL}, which can be described in three steps.
\begin{itemize}
    \item \textbf{Step 1.} Each client trains their local model and updates model parameters during each training round. Then, each client passes the updated parameters to a central server.
    \item \textbf{Step 2.} The global model aggregates the updated parameters from all local clients and updates its parameters accordingly in the central server.
    \item \textbf{Step 3.} The updated global model parameters are then broadcast to each local client, and these three steps are iterated for multiple rounds until the convergence is reached.
\end{itemize}

\subsection{Differential privacy}
DP introduces noise into the raw dataset so that it provides statistical guarantees against the information a malicious adversary may infer from the output of a randomized algorithm \cite{dwork2014algorithmic}.
\begin{definition}[\textbf{Differential Privacy \cite{abadi2016deep}}]
A random algorithm $\mathcal M$ is compliant with ($\epsilon$, $\delta$)-DP if for any two neighboring input datasets $L, L'$ and for any subset of outputs/events $\pmb{S} \subseteq Rang(\mathcal M)$,
\begin{equation}\label{eq4}
\operatorname{Pr}[\mathcal M(L) \in \pmb{S}]\leq e^\epsilon \operatorname{Pr}[\mathcal M(L') \in \pmb{S}] + \delta.
\end{equation}
\end{definition}
In the above equation, $\epsilon$ is the privacy budget/loss, which is inversely proportional to the privacy level. $\delta$ is the probability that the upper privacy bound is broken, i.e., the occurrence of a bad event. It is a plain $\epsilon$-DP when $\delta$ equals 0.

For a real-valued function $\mathcal F$, a common exemplification is to calibrate an additive zero-mean Laplacian or Gaussian noise mechanism to the sensitivity of $\mathcal F$, which can be denoted as 
\begin{equation}
    \Delta \mathcal F = \max_{L,L'} \left \| \mathcal{F}(L)-\mathcal{F}(L') \right \| _1.
\end{equation} Depending on whether a single record is included or excluded, the sensitivity $\Delta \mathcal F$ measures the maximum change in output.

The Gaussian mechanism adds Gaussian noises to $\mathcal F$ to satisfy ($\epsilon, \delta$)-DP: $\forall \delta \in (0, 1)$, the noise is denoted by $\mathcal N(0, {\Delta \mathcal F}^2 \cdot \sigma^2)$, and we have
\begin{equation}\label{eq5}
\mathcal M(L) = \mathcal F(L) + \mathcal N(0, {\Delta \mathcal F}^2 \cdot \sigma^2),
\end{equation}
where $\Delta \mathcal F \cdot \sigma$ is the standard deviation, and $\sigma\ge\frac{\sqrt{2\ln(1.25/\delta)}}{\epsilon}$. 

\section{DP$^2$-NILM Framework}\label{DP2NILMFramework}

\subsection{Overview of DP$^2$-NILM}
\begin{figure}[!htbp]
	\centering
	\includegraphics[width=0.9\textwidth]{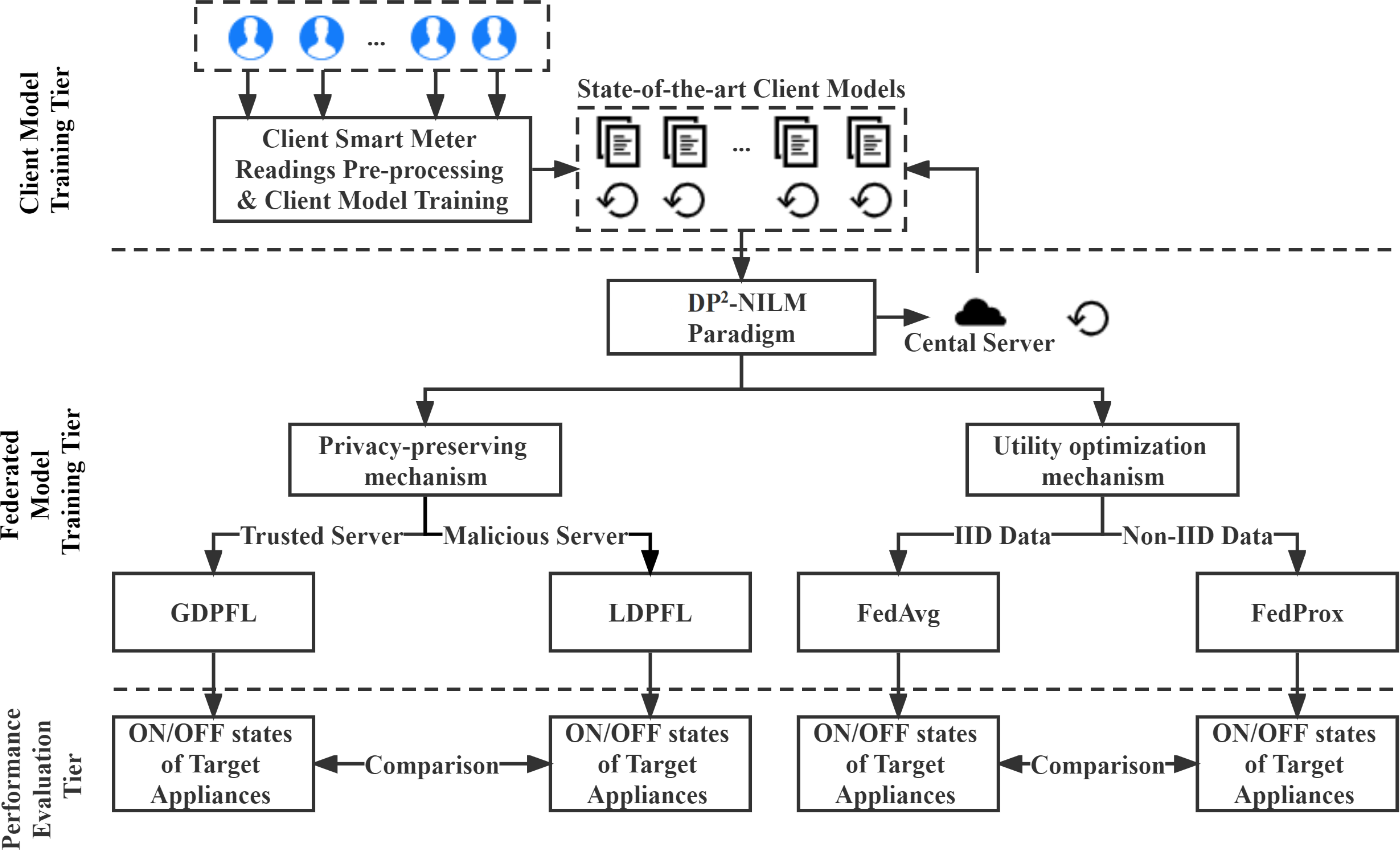}
	\caption{The workflow of proposed DP$^2$-NILM framework.}
	\label{FigDP2NILMFramework}
\end{figure}

The key objective of our DP$^2$-NILM framework is to train different federated learning models focusing on various enhancement schemes, i.e., the utility optimization schemes and the privacy-preserving schemes, for real-world NILM application scenarios. We further stress that the DP$^2$-NILM framework is easily extensible to incorporate various state-of-the-art DNN models and datasets. As presented in Figure \ref{FigDP2NILMFramework}, the whole workflow of the DP$^2$-NILM framework contains three tiers.

\begin{itemize}
    \item \textbf{Client Model Training Tier.} In this tier, smart meter readings from the client side are preprocessed into standard formats for the federated pipeline. The client can either specify their privacy-preserving or the data heterogeneity optimization requirements. After preprocessing, each client trains their data based on a state-of-the-art DNN model, which will be introduced in Section \ref{DLM}, and then upload their parameters through the DP$^2$-NILM paradigm.
    \item \textbf{Federated Model Training Tier.} This tier is the key part of the DP$^2$-NILM framework. Based on the special requirement from the client model training tier, the DP$^2$-NILM assigns different federated learning mechanisms to each client. For example, a client-side requires a strict privacy-preserving mechanism to protect its sensitive data. After receiving this request, DP$^2$-NILM will deliver a high-level privacy-preserving paradigm, the local differential privacy federated learning (Section \ref{LDPFL}), to train the FL model based on the typical FL training steps.
    
    During FL training, the objective for $N$ clients can be described as an optimization problem:

    \begin{equation}\label{FLobj}
    \begin{aligned}
    \underset{w_g}{\min} \mathcal L_{g}(w_g) =& \frac{1}{\left| L \right|}\sum_{n=1}^{N}\left|L^n\right|\cdot \mathcal{L}_{c}^n(w_c) \\
    \text { where } \quad & \mathcal{L}_{c}^n(w_c) = \frac{1}{\left| L^n \right|}\sum_{i \in L^n} \mathcal{L}_{i}(w_i),\\
    & \forall L^{n} \in L, n \in\{1,2,\ldots, N\}
    \end{aligned}
    \end{equation}
    where $L_g(w_g)$ is the loss of the global model, $\mathcal{L}_c^n(w_c)$ is the loss of the $n$-th local client model, and $\mathcal{L}_i(w_i)$ is the loss of a single smart meter reading. Each household $n \in \left\{ 1,2,\ldots,N \right\}$, and generates its private smart meter readings $L^n = \left\{ (l^n, s^n), (l^n, s^n),\ldots,(l^n, s^n) \right\}$, where $l^n$ is the aggregated load consumption of the target appliances, and $s^n$ is the corresponding states (ON/OFF) set of these appliances. 
    
    The most commonly used optimization algorithms for FL is the FedAvg \cite{dianxin1-Communication}. Based on the FedAvg, two subsequent research streams for enhancing the FL paradigm have been proposed, i.e., the utility optimization schemes and the privacy preserving schemes. Following this development, the DP$^2$-NILM framework uses the FedAvg as the baseline to include the above two enhancing schemes. Specifically, the DP$^2$-NILM adopts the FedAvg and the FedProx to optimize the model utility for FL-based NILM. Furthermore, studies \cite{wei2020federated, wei2020federated} have provided clear theoretical foundations for GDPFL and LDPFL based on the FedAvg, and hence the DP$^2$-NILM develops GDPFL and LDPFL in privacy-preserving schemes based on FedAvg.
    \item \textbf{Performance Evaluation Tier.} We designed different model training paradigms for different NILM application scenarios based on three real-world smart meter datasets, and the model performance of each scenario are evaluated and validated in this tier.
\end{itemize}

\subsection{State-of-the-art NILM client model} \label{DLM}
\begin{figure}[!htbp]
	\centering
	\includegraphics[width=0.9\textwidth]{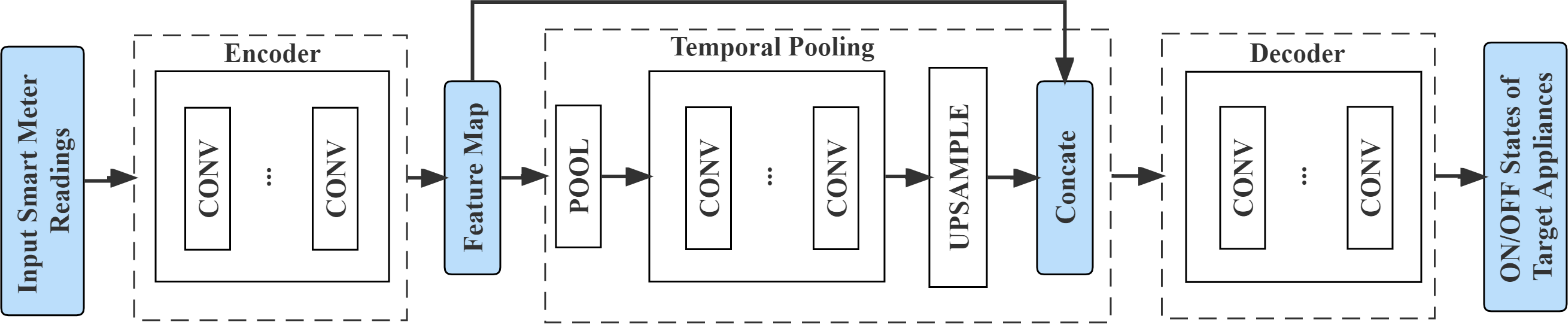}
	\caption{The overall layout of the deep learning model for NILM.}
	\label{DP2NILM_DNN}
\end{figure}

We introduce a state-of-the-art deep learning architecture, i.e., the pyramid scene parsing network (PSPNet) \cite{PSPNet}, to enhance the performance of DP$^2$-NILM in both the local client model training and the central server global model training, which was originally used for image semantic segmentation. The selection of this particular architecture is motivated by its potentially promising performance in learning the inherent signatures of appliances as demonstrated in \cite{massidda2020non}. We further adjusted the PSPNet model for the NILM task, and the training structure of the adjusted PSPNet is shown in Figure \ref{DP2NILM_DNN}.

We provide a detailed description of the adjusted PSPNet model, which consists of three modules: the encoder, the temporal pooling module, and the decoder.
\begin{itemize}
    \item \textbf{Encoder.} The input of the encoder is the household aggregated load consumption of the target appliances over a 1-hour interval (the consumption datasets were resampled to 30 seconds). The encoder makes up of four modules, each of which is alternated by a max pool layer except for the last block. The encoder increases the output features from a single aggregation value to 256, while paying the price of decreasing the time signal resolution by 10 times.
    \item \textbf{Temporal Pooling.} The temporal pooling consists of four average pooling modules, filter sizes of which are decreased from the whole size of the input signal to one-sixth of it. After going through a convolutional layer, the feature dimension of the input is reduced to a quarter of its original size, and the acquired feature maps were upsampled to the size of the input time signals. Then the upsampled feature maps (shallow features) are concatenated with the original input signal (deep features) from the temporal pooling to get the final feature maps. The fusion of the deep and shallow features of the temporal pool could enable this block to get contextual information fed into the decoder.
    \item \textbf{Decoder.} The decoder receives the output from the temporal pooling block and passes it to a convolutional layer to recover the time signal resolution. Then the output is fed into the final convolutional layer to produce the final appliance-level load disaggregation.
    
\end{itemize}
\section{Utility Optimization of DP$^2$-NILM}\label{UoDP2NILM}
Recall the FL optimization objective (equation \ref{FLobj}), the DP$^2$-NILM framework considers two utility optimization shcemes, the FedAvg-NILM and the FedProx-NILM, to achieve this goal.
\subsection{FedAvg-NILM}
Algorithm \ref{AgFedAvg} depicts the steps of FedAvg-NILM.
\begin{algorithm}[!htbp]\label{AgFedAvg}  
	\LinesNumbered 
	\scriptsize
	\caption{FedAvg-NILM} 
	
    \textbf{Central Server Execution:}\\
	Initialize the global model parameters $w_G$\\
	\For {each global round $r\leq{R}$} 
	{
		
		\For{each client $n \in \left\{1,2,\ldots,N\right\}$ \textbf{in parallel}}{
			$w_{r+1}^n \leftarrow HouseholdsUpdate(w_{r}^n)$\
		}
		$w_{r+1} \leftarrow \frac{\sum_{n=1}^Nw_{r}^n}{N}$\
	}
	Broadcast the global model to all clients
	
	\textbf{Smart Meter Client Execution:}\\
	\textbf{procedure} $HouseholdsUpdate(w_{r}^n)$:\\
	Split $L_n$ into batches of size $B_L$\; 
	
	\For{ each local client epoch ${e}\le E$}{
		\For{each batch of $L_n$}{
			$w^n \leftarrow w^n - \eta \nabla\mathcal{L}(w^n)$\
		}
	}
	Upload $w^n$ to the central server\\
\end{algorithm}
The FedAvg \cite{dianxin1-Communication} allows the smart meter clients to train their local DNN models iteratively using the same learning rate and the number of epochs before uploading the updated model weights to the central server. For each global round (line 3), every smart meter client receives a copy of the global model and trains its local DNN models with its own private smart meter readings for multiple epochs using $w^n \leftarrow w^n - \eta \nabla\mathcal{L}(w^n)$\ (line 13-17), where $\eta$ is the learning rate. After this, the local clients upload their updated local model weights $w^n$ to the central server (line 18). Then, the central server updates the global model by averaging the uploaded weights from the smart meter clients (line 7) and broadcasts the updated global model to all clients (line 9).

An advantage of FedAvg-NILM is that a well-trained FedAvg-NILM model can outperform a single local NILM model while maintaining data privacy. Moreover, FedAvg has been proved to be efficient in reducing the communication overhead between the local clients and the global server \cite{dianxin1-Communication}.

Nevertheless, the FedAvg only performs effectively under the premise of all the local clients utilize the similar initialization, and it has been shown that heterogeneity of data impedes the convergence of FedAvg \cite{li2020federated}. On the other hand, in the real-world NILM tasks, smart meter clients often exhibit diverse appliance usage patterns, making the local client models easy to deviate from the global model, thereby reducing the overall performance. 
\subsection{FedProx-NILM}
It is likely that data from smart meters are heterogeneous since they are collected under various contexts (e.g., across different countries) and are affected by diverse client behaviours leading to heterogeneous load usage distributions. Our DP$^2$-NILM framework is efficient for guaranteeing the convergence of the FL model in heterogeneity settings, i.e., the non-IID data settings, by incorporating FedProx \cite{li2020federated} as an extension of the utility optimization scheme.

\begin{algorithm}[!htbp]\label{AgFedProx}  
	\LinesNumbered 
	\scriptsize
	\caption{FedProx$_{mod}$-NILM} 
	
    \textbf{Central Server Execution:}\\
    (// Same central server execution steps as the FedAvg-NILM)\\
	Broadcast the global model to all clients
	
	\textbf{Smart Meter Client Execution:}\\
	\textbf{procedure} $HouseholdsUpdate(w_{r}^n)$:\\
	Split $L_n$ into batches of size $B_L$\ 
	
	\For{ each local client epoch ${e}\le E$}{
		\For{each batch of $L_n$}{
        	$\nabla\mathcal{L}_{prox}(w^n) \leftarrow \nabla\mathcal{L}(w^n) + \mu(w^n-w_r^n)$\ \\
        	$w^n \leftarrow w^n - \eta \nabla\mathcal{L}_{prox}(w^n)$ \
		}
	}
	Upload $w^n$ to the central server\\
\end{algorithm}
Algorithm \ref{AgFedProx} depicts the steps of FedProx-NILM. The central server executes the same steps as in the FedAvg-NILM. However, a proximal term $\mu(w^n-w_r^n)$ is added to update the local model of smart meter clients (line 9), which keeps local updates from deviating too much from the initial global model. When $\mu=0$, the FedProx-NILM will produce the same results as the FedAvg-NILM.

Specifically, in the typical FedProx training paradigm, there is an inexact minimizer adjusting the local epoch of each client to reduce the negative impact of the system heterogeneous, which is defined as follows.
\begin{definition}[$\gamma$-inexact Solution \cite{li2020federated}]
The $w^*$ is a $\gamma$-inexact minimizer solution for the optimization objective in equation \ref{FLobj} if $\left \|w^*-w_r^n\right \| \leq \gamma \left \| w_r^n-w_{r-1}^n \right \|$, where $\gamma \in [0, 1)$.
\end{definition}

The $\gamma$-inexact minimizer solution considers adjusting the local computation and the global communication overhead based on the number of local model epochs performed by the clients. In our framework, we hypothesize that most smart meter clients are available and capable of completing a certain number of local epochs whereas for the very few stragglers, their destabilized training environment may produce models that contribute little to the FL global model. Therefore, we adjusted the FedProx to make it more efficient in the DP$^2$-NILM framework by utilizing the proximal term $\mu(w^n-w_r^n)$ with the exact minimizer solution $w_r^n$ rather than the inexact one. 

\section{Privacy-preserving of DP$^2$-NILM}\label{PpDP2NILM}
The DP$^2$-NILM considers privacy-preserving mechanisms at two different levels to suit various privacy requirements from smart meter clients, i.e., the global differential privacy federated learning and the local differential privacy federated learning.
\subsection{Global differential privacy federated learning NILM}\label{GDPFL}
In the DP$^2$-NILM paradigm, if a client sends out the privacy requirement and meanwhile trusts the central server, the GDPFL-NILM will be utilized for this client. Although there must be a certain degree of trust in the central server, this presumption is significantly less stringent than granting the server access to the data. Algorithm \ref{AgGDPFL} details GDPFL-NILM scheme in the DP$^2$-NILM.  
\begin{algorithm}[!htbp]\label{AgGDPFL}  
	\LinesNumbered 
	\scriptsize
	\caption{GDPFL-NILM} 
	
    \textbf{Central Server Execution:}\\
	Initialize the global model parameters $w_G$\\
	\For {each global round $r\leq{R}$} 
	{
		Compute privacy cost:
		$\hat{\epsilon}_{r} \leftarrow PrivacyAccount(\delta, \sigma)$\;
		\If {$\hat{\epsilon}_{r} > \epsilon_{r} $}{
  	\textbf{return} $w_{r}$\\}
  	    \Else{
		\For{each client $n \in \left\{1,2,\ldots,N\right\}$ \textbf{in parallel}}{
			$w_{r+1}^n \leftarrow HouseholdsUpdate(w_{r}^n)$\
		}
		$w_{r+1} \leftarrow \frac{\sum_{n=1}^Nw_{r}^n}{N} +  \mathcal N(0, {\Delta \mathcal F}^2 \cdot \sigma^2)$\
	}
	}
	Broadcast the global model to all clients
	
	\textbf{Smart Meter Client Execution:}\\
    (// Same smart meter client execution steps as the FedAvg-NILM)\\
	Upload $w^n$ to the central server\\
\end{algorithm}

In GDPFL-NILM, the smart meter client execution steps are the same as in FedAvg-NILM. The central server guarantees participant-level privacy by perturbing the model weights aggregation, i.e., adding Gaussian noise $\mathcal N(0, {\Delta \mathcal F}^2 \cdot \sigma^2)$ to the aggregated results (line 12). Moreover, to ensure the $(\epsilon, \delta)$-GDP, after each global round, the algorithm $PrivacyAccount()$ calculates the accumulated privacy budget (line 4), and if it exceeds the overall budget $\epsilon$, the global training iteration will be stopped (line 6). 

\subsection{Local differential privacy federated learning NILM}\label{LDPFL}
In the LDPFL \cite{abadi2016deep}, smart meter clients apply noise on the updated local model weights before uploading them to the central server. The LDPFL-NILM scheme in DP$^2$-NILM is presented in Algorithm \ref{AgLDPFL}.
\begin{algorithm}[t]\label{AgLDPFL} 
	\LinesNumbered 
	\scriptsize
	\caption{LDPFL-NILM} 
	
    \textbf{Central Server Execution:}\\
    (// Same central server execution steps as the FedAvg-NILM)
     \\
		
	Broadcast the global model to all clients
	
	\textbf{Smart Meter Client Execution:}\\
	\textbf{procedure} $HouseholdsUpdate(w_{r}^n)$:\\
	Split $L_n$ into batches of size $B_L$\ 
	
	\For{ each local client epoch ${e}\le E$}{
		\For{each batch of $L_n$}{
        	$\nabla\mathcal{L}_{ldp}(w^n) \leftarrow \nabla\mathcal{L}(w^n) + \mathcal N(0, \frac{{\Delta \mathcal F}^2 \cdot \sigma^2}{N})$\ \\
        	$w^n \leftarrow w^n - \eta \nabla\mathcal{L}_{ldp}(w^n)$ \
		}
	}
	Upload $w^n$ to the central server\\
\end{algorithm}

The central server updating process in the LDPFL-NILM is the same as in FedAvg-NILM. On the other hand, the smart meter clients guarantee their own privacy by perturbing the updated local model weights, i.e., adding Gaussian noise $\mathcal N(0, \frac{{\Delta \mathcal F}^2 \cdot \sigma^2}{N})$ to the updated model weights (line 9). The LDPFL-NILM provides better privacy notion than the GDPFL-NILM, and therefore it is suitable for clients who requires strict data privacy-preserving discipline. 

\section{Performance Evaluation}\label{PE}
In this section, we use real-world smart meter datasets to evaluate the proposed DP$^2$-NILM framework. The datasets and the evaluation criteria are firstly introduced. Then, the performance of the FL setting in DP$^2$-NILM is compared with the Local-NILM models trained on individual household datasets and the Centralized-NILM model trained on aggregated household datasets. After this, we examine the utility optimization schemes in DP$^2$-NILM paradigm. Finally, based on the FedAvg, two privacy-preserving schemes, i.e., the GDPFL-NILM and the LDPFL-NILM, are compared in terms of the trade-off between model utility and privacy.
\subsection{Experimental settings}
\subsubsection{Dataset and preprocessing} 

We used three real-world smart meter datasets to evaluate the DP$^2$-NILM framework, including UKDALE \cite{yuan54}, REDD \cite{kolter2011redd} and REFIT \cite{murray2017electrical}. 
\begin{itemize}
    \item UKDALE: The U.K. Domestic Appliance Level Electricity (UKDALE) dataset contains five buildings in the U.K. between 2013 and 2015 with 1$s$ sampling period for mains and 6$s$ sampling period for appliances.
    \item REDD: The Reference Energy Disaggregation Dataset (REDD) consists of six buildings in the U.S. from 3 to 19 days, and the sampling periods for mains and appliances are 1$s$ and 6$s$, respectively. 
    \item REFIT: The REFIT dataset contains 20 buildings in the U.K. between 2013 and 2015 with 8$s$ sampling period for mains and appliances.
\end{itemize}

Three appliances (fridge, dishwasher, washing machine) are selected as our target appliances for comparison purpose. 80\% records from each smart meter client were selected as the training set, followed by a 10\%  for validation, and 10\%  for testing. The distribution of the selected buildings are listed in Table \ref{tabData}.
\begin{table}[H]
	\centering
		\caption{Distribution of the selected datasets. D: Days}
		\begin{center}
		\resizebox{0.9\textwidth}{!}{
			\begin{tabular}{cccccc}
\hline
\textbf{Datasets}                & \textbf{Building} & \textbf{Period}                   & \textbf{Total (D)} & \textbf{Training (D)} & \begin{tabular}[c]{@{}c@{}}\textbf{Validation}\\ \textbf{/Testing (D)}\end{tabular} \\ \hline
\multirow{3}{*}{UKDALE} & 1        & 2013-04-12 to 2017-04-25 & 1475      & 1180         & 147.5                                                             \\
                        & 2        & 2013-05-22 to 2013-10-03 & 135       & 108          & 13.5                                                              \\
                        & 5        & 2014-06-29 to 2014-09-01 & 65        & 52           & 6.5                                                               \\ \hline
\multirow{3}{*}{REDD}   & 1        & 2011-04-19 to 2011-05-19 & 31        & 24.8         & 3.1                                                               \\
                        & 2        & 2011-04-18 to 2011-05-21 & 34        & 27.2         & 3.4                                                               \\
                        & 3        & 2011-04-17 to 2011-05-30 & 44        & 35.2         & 4.4                                                               \\ \hline
\multirow{3}{*}{REFIT}  & 2        & 2013-09-18 to 2015-05-27 & 617       & 493.6        & 61.7                                                              \\
                        & 5        & 2013-09-27 to 2015-07-05 & 647       & 517.6        & 64.7                                                              \\
                        & 9        & 2013-12-18 to 2015-07-07 & 567       & 453.6        & 56.7                                                              \\ \hline
\end{tabular}
			}
			\label{tabData}
	\end{center}
\end{table}

TABLE \ref{tabThresholds} gives the relevant thresholds used in data preprocessing. We firstly filtered out the abnormal load consumption by the max power \cite{kelly}, and then down-sampled the load consumption of all the 9 households from 6$s$ to 30$s$ through averaging. After this, the resampled data were normalized by subtracting the mean and dividing a constant load value 2000 W following \cite{kelly}. Then the state series of each target appliance was derived from activation-time thresholding \cite{kelly} as the input to feed into the DP$^2$-NILM.

\begin{table}[H]
	\centering
		\caption{Relevant threshold information}
		\begin{center}
		\resizebox{0.7\textwidth}{!}{
			\begin{tabular}{cccc}
				\hline
				& \textbf{Fridge} & \textbf{Dishwasher} & \textbf{Washing Machine} \\
				\hline
				\textbf{Max power (W)}         & 300    & 2500       & 2500            \\ \hline
				\textbf{Power threshold (W)}   & 50     & 20         & 20              \\ \hline
				\textbf{Min. ON duration (s)}  & 1      & 60         & 60              \\ \hline
				\textbf{Min. OFF duration (s)} & 0      & 60         & 5               \\ \hline

			\end{tabular}
			}
			\label{tabThresholds}
	\end{center}
\end{table}

Furthermore, parameters used in the DP$^2$-NILM framework are listed in TABLE \ref{tabParameter}. We used TensorFlow to train the DP$^2$-NILM framework. To keep the comparison fair, all the models in our experiment use the same DNN architecture described in Section \ref{DP2NILM_DNN}. For all the FL models in DP$^2$-NILM, each global training round consists of eight local epochs, allowing the clients to take reasonable learning steps before central server aggregation. For the privacy-preserving scheme, we vary the privacy budget $\epsilon$ between 4 and 12 while keeping $\delta=10^{-5}$, and report the performance and attack success risk, i.e., the Accuracy and the ASR. The choice of $\delta = 10^{-5}$ satisfies the requirement that $\delta$ should be smaller than the inverse of the training data size \cite{abadi2016deep}. To bound the sensitivity $\Delta {\mathcal F}^2$ of the gradients, clipping is required, which is a computationally efficient and common practice in deep learning. With the TensorFlow Privacy framework, we implemented the batch clipping with a threshold of 4. Further, for the listed parameters, we note that an additional parameter tuning step may improve the final model performance, however at the cost of massive computational resources. 

\renewcommand\arraystretch{0.9}
\begin{table}[H]
	\centering
		\caption{Parameters used in the DP$^2$-NILM framework}
		\begin{center}
		\resizebox{0.5\textwidth}{!}{
			\begin{tabular}{ll}
\hline
\textbf{Parameters}                       & \textbf{Value}               \\ \hline
Batch size                       & 32                  \\ \hline
Global rounds                    & 10
\\ \hline
Local epochs                     & 8                   \\ \hline
Number of clients                & 9                   \\ \hline
Proximal parameter $\mu$         & 0.01                \\ \hline
Privacy budget $\epsilon$        & {[}4, 8, 12{]}
\\ \hline
Privacy relaxation term $\delta$ & 10$^{-5}$  \\ \hline
Gradient clipping threshold      & 4                   \\ \hline
Learning rate $\eta$             & 10$^{-4}$           \\ \hline
Activation function              & ReLU                \\ \hline
Dropout probability              & 0.1                 \\ \hline
Momentum                         & 0.5                 \\ \hline
Optimizer                        & SGD                 \\ \hline
\end{tabular}
			}
			\label{tabParameter}
	\end{center}
\end{table}

\subsubsection{Evaluation criteria} \label{EC}
Four evaluation metrics are used to assess the model performance of the DP$^2$-NILM framework. Denote true positive as TP, true negative as TN, false positive as FP, and false negative as FN, the evaluation metrics can be defined as follows:
\begin{equation}\label{eq13}
Precision = \frac{TP}{TP+FP}
\end{equation}

\begin{equation}\label{eq14}
Recall = \frac{TP}{TP+FN} \\
\end{equation}

\begin{equation}\label{eq15}
Accuracy = \frac{TP+TN}{TP+TN+FP+FN}\\
\end{equation}

\begin{equation}\label{eq16}
F_1 = 2 \times \frac{Precision\times Recall}{Precision + Recall}
\end{equation}
where the precision represents the proportion of $TP$s to all the data sequences classified to the ON state. Recall denotes the ratio of $TP$s to all data sequences that are actually in the ON state. Accuracy reflects the ratio of all correctly identified samples to all the smart meter data sequences. $F_1$ is defined as a weight average representation for the precision and the recall within the range of $[0, 1]$. An $F_1$ close to 1 indicates that the classification results for the target appliances are better. 

Moreover, to measure the privacy risk for the privacy-preserving DP$^2$-NILM models, we utilized the member inference attack metric defined in \cite{yeom2018privacy}. To test the membership of an input record, this attack mechanism evaluates the loss of the uploaded local model parameters and then classifies it as a member if the loss is smaller than the average training loss. The attack success risk can be calculated as
\begin{equation}
    ASR = TPR-FPR,
\end{equation}
where $TPR = \frac{TP}{TP+FN}$ denotes the TP rate, and $FPR = \frac{FP}{FP+TN}$ represents the FP rate. 
%

\subsection{Evaluation on the baseline model of DP$^2$-NILM}
This subsection evaluates the FL setting of DP$^2$-NILM. There are three different model settings in this subsection:
\begin{itemize}
    \item Local-NILM models: The Local-NILM models are trained on 9 household datasets separately. This setting eliminates the need for data sharing with the central server, but at the expense of having to update all of the 9 models separately.
    \item Centralized-NILM model: The Centralized-NILM model is trained on aggregated datasets from all the 9 households, which requires raw data sharing from the smart meter clients.
    \item FL-setting of DP$^2$-NILM (FedAvg-NILM): The FL-setting of DP$^2$-NILM utilizes FedAvg as the optimization method and trained on all the 9 households without any exchange of the raw smart meter data. The FL model trained based on FedAvg in DP$^2$-NILM will be used later as the baseline for evaluating two schemes in the proposed framework.
\end{itemize}

For the Local-NILM models and the Centralized-NILM model, the epochs are set to 80 to achieve the final convergence. For comparison purposes, Table \ref{tabAvgScores} lists the average performance scores of the Local-NILM models, the Centralized-NILM model and the FL-setting of DP$^2$-NILM (FedAvg-NILM).  

\renewcommand\arraystretch{1.2}
\begin{table}[H]
	\centering
		\caption{Average performance scores of the Local-NILM models, the Centralized-NILM model, and the FL-setting of DP$^2$-NILM for 9 households}
		\begin{center}
		\resizebox{\textwidth}{!}{
\begin{tabular}{ccccccccccccc}
\hline
                                & \multicolumn{4}{c}{\textbf{Fridge}}   & \multicolumn{4}{c}{\textbf{Dishwasher}} & \multicolumn{4}{c}{\textbf{Washing Machine}} \\ \cline{2-13} 
                                & Accuracy & F$_1$ & Precision & Recall & Accuracy  & F$_1$  & Precision  & Recall & Accuracy   & F$_1$   & Precision   & Recall  \\ \hline
\textbf{Local-NILM}      & 0.83     & 0.79  & 0.82      & 0.74   & 0.98      & 0.88   & 0.86       & 0.83   & 0.99       & 0.78    & 0.89        & 0.70    \\ \hline
\textbf{Centralized-NILM} & 0.86     & 0.80  & 0.79      & 0.81   & 0.97      & 0.70   & 0.87       & 0.59   & 0.97       & 0.66    & 0.71        & 0.62    \\ \hline
\textbf{FedAvg-NILM}            & 0.65     & 0.63  & 0.50      & 0.85   & 0.97      & 0.75   & 0.92       & 0.64   & 0.98       & 0.71    & 0.83        & 0.62    \\ \hline
\end{tabular}
			}
			\label{tabAvgScores}
	\end{center}
\end{table}

A comparison of FedAvg-NILM with Local-NILM models examines the performance of federated learning strategies in capturing diversities among clients. Moreover, comparing FedAvg-NILM to the Centralized-NILM model evaluates the overall performance of the common utility FL model. From the results, we can see that for each appliance, all models achieved satisfactory results on the dishwasher and washing machine, and reasonable results on the fridge. Note that the FedAvg-NILM achieved the same accuracy score and higher F$_1$, precision, and recall scores on dishwashers and washing machines compared with the centralized-NILM model. For fridge, as it consumes relatively low power compared with other appliances, it is likely to be learned with less evident signature during model training and such consumption can easily be omitted as unidentified load noise in the FL paradigm. Overall, we can conclude the FedAvg-NILM model in the DP$^2$-NILM framework works well and its performance can be used as the baseline for further evaluations.

We also observed that with more global rounds, the FedAvg-NILM may achieve more satisfying performances. For example, we have set 100 global rounds for the FedAvg-NILM, and the final obtained average accuracy for the fridge, dishwasher, and washing machine were 0.89, 0.99, and 0.99, respectively, which are even better than the centralized NILM model. However, parameter tuning in FL remains a challenge as a result of the distributed environment and the associated computational overhead \cite{kairouz2021advances}. We further stress that fixed global and local training rounds enable efficient, fair, and comparable evaluations in our DP$^2$-NILM framework.

\subsection{Evaluations on utility optimization of DP$^2$-NILM} 
\begin{figure}[H]
\centering
    \subfigure[UKDALE House 1]{
\includegraphics[width=3.3cm]{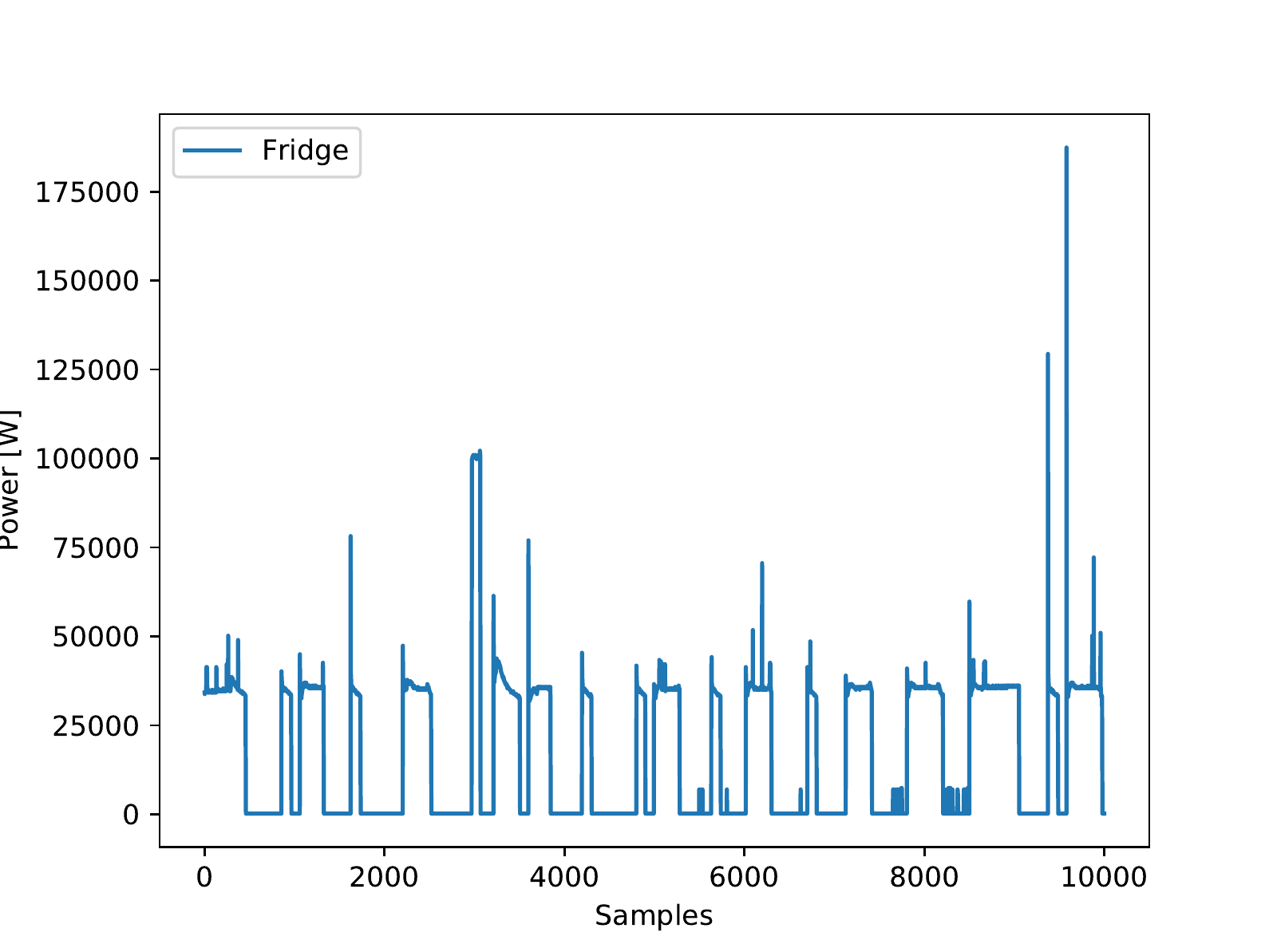}
}
\quad
\subfigure[UKDALE House 2]{
\includegraphics[width=3.3cm]{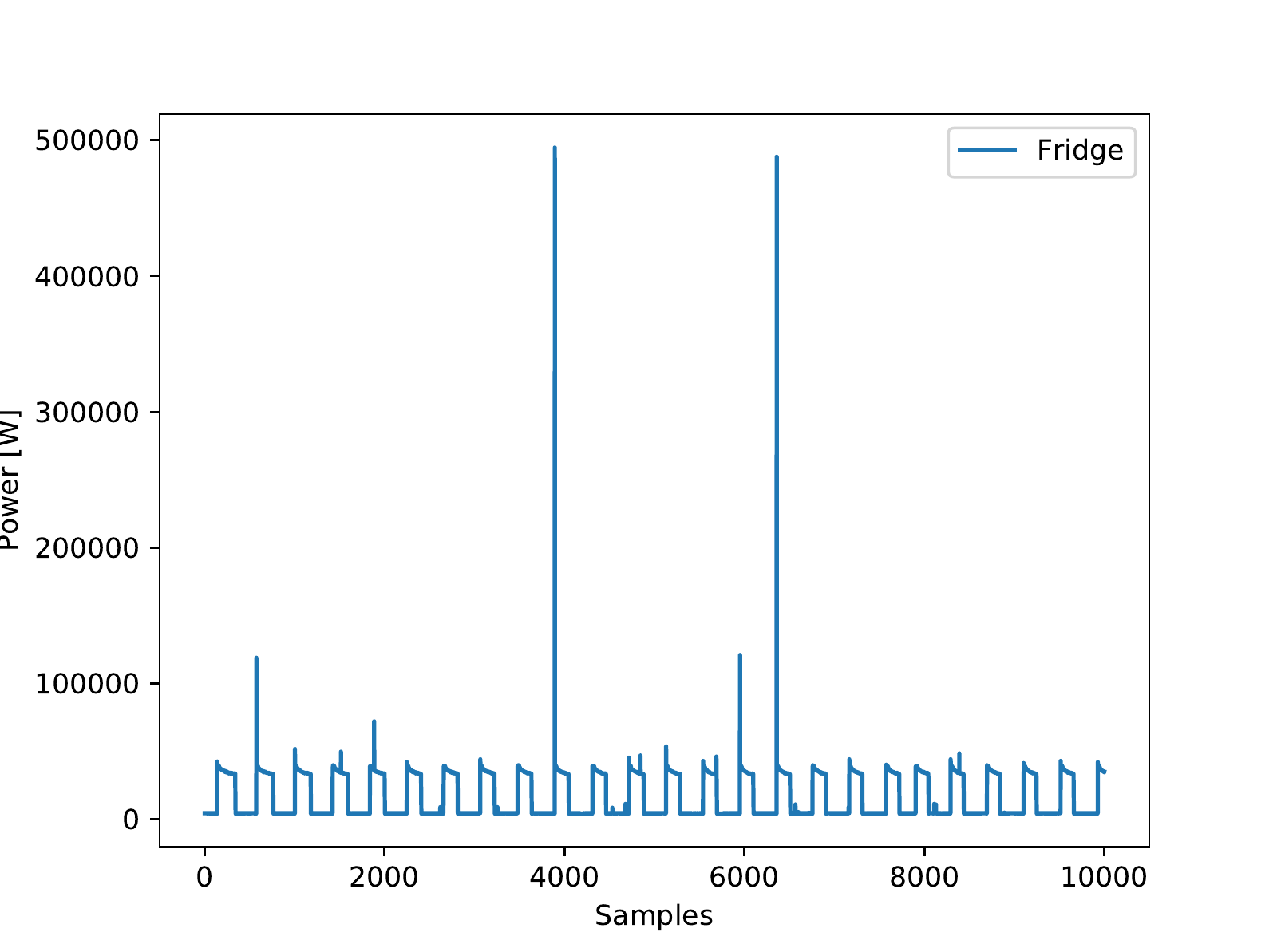}
}
\quad
\subfigure[UKDALE House 5]{
\includegraphics[width=3.3cm]{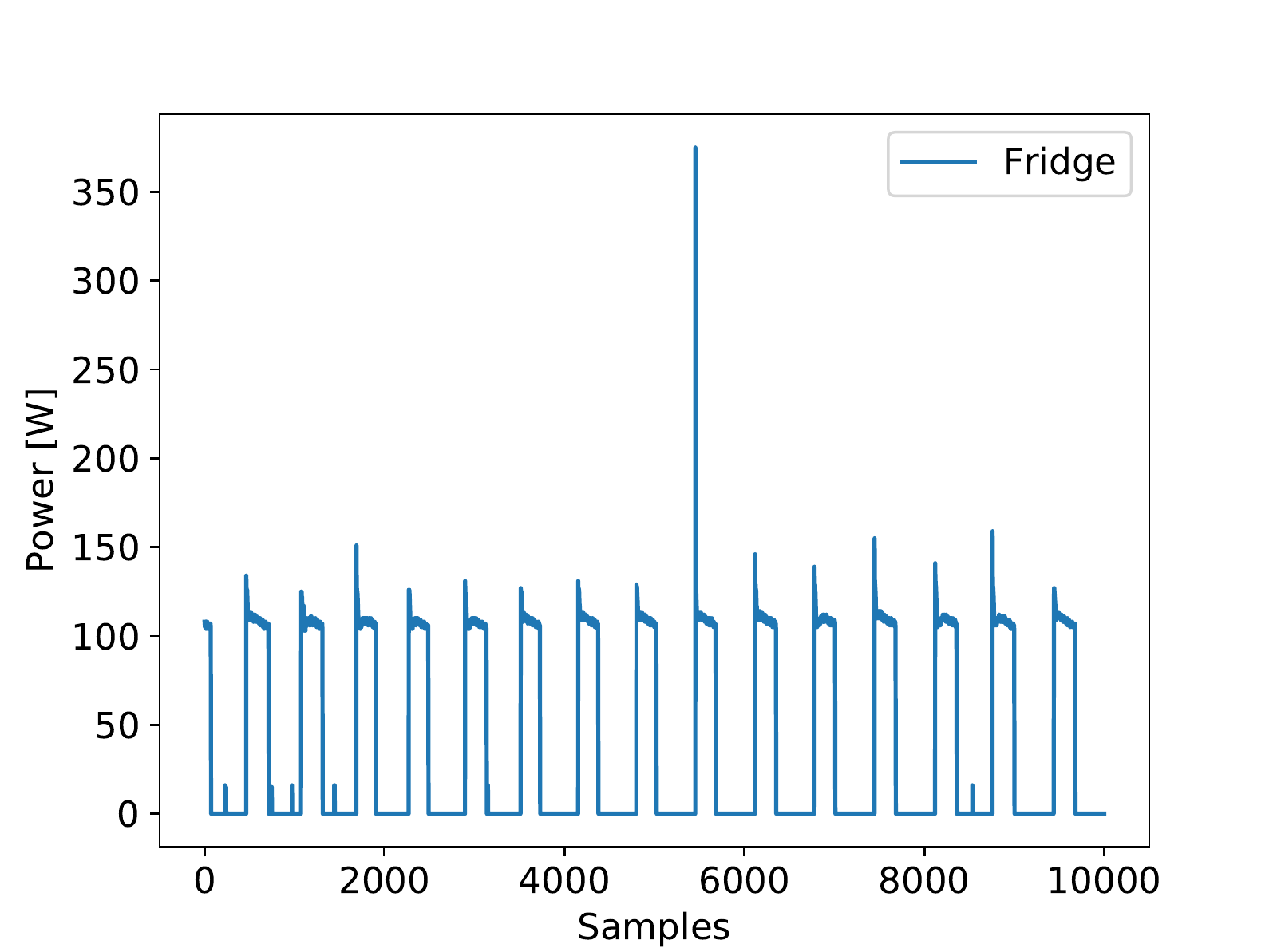}
}
\quad
\subfigure[REDD House 1]{
\includegraphics[width=3.3cm]{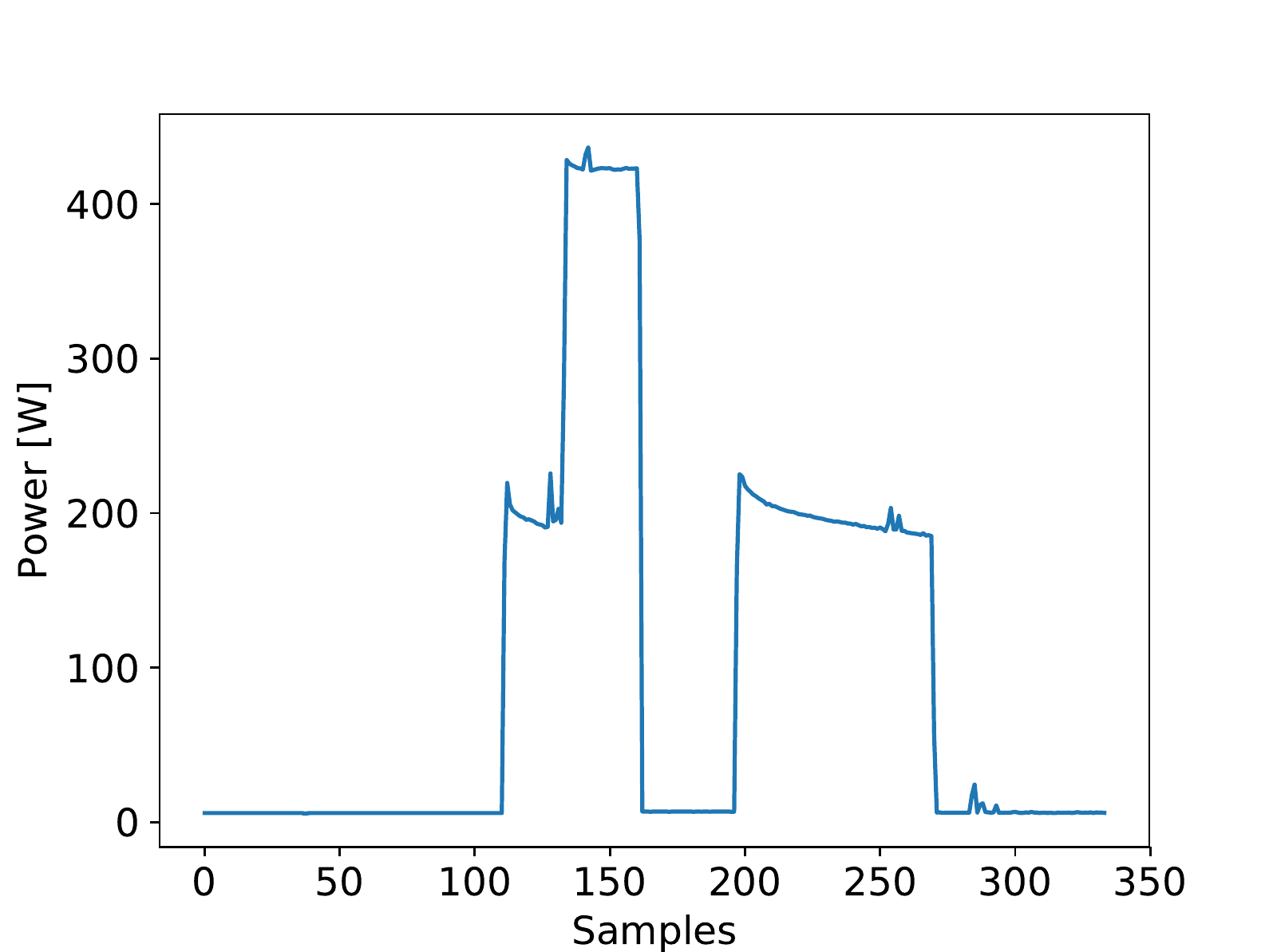}
}
\quad
\subfigure[REDD House 2]{
\includegraphics[width=3.3cm]{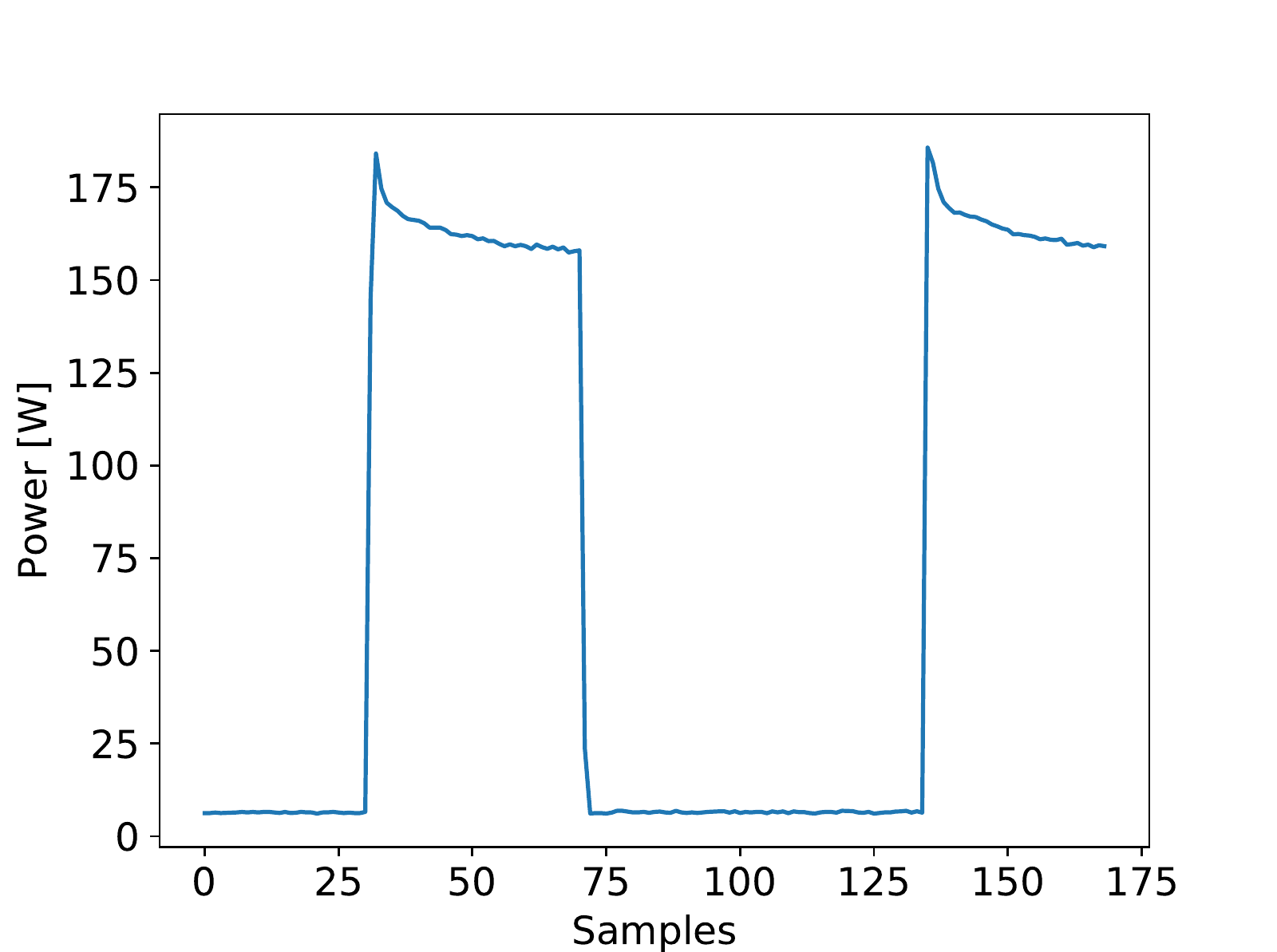}
}
\quad
\subfigure[REDD House 3]{
\includegraphics[width=3.3cm]{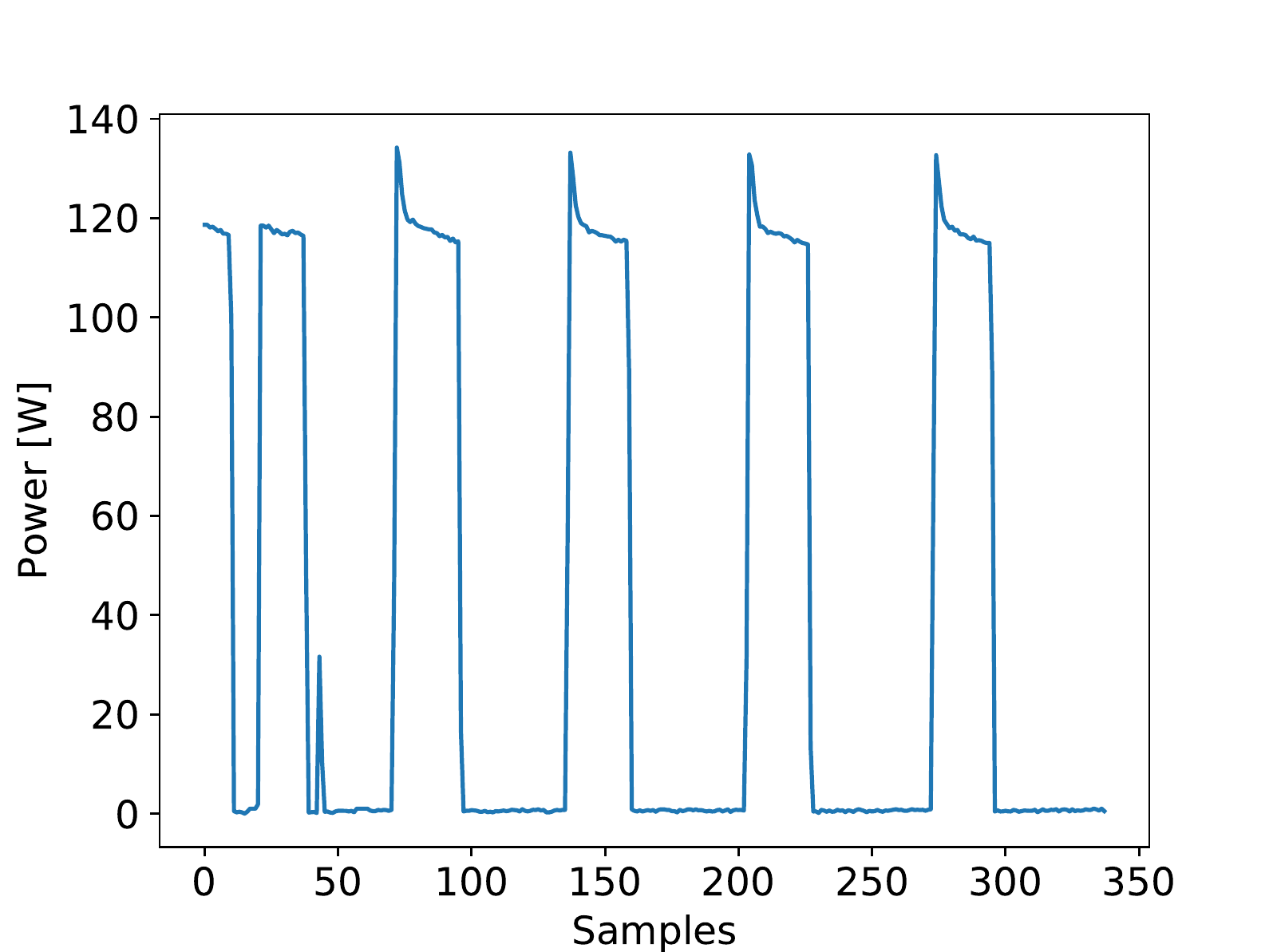}
}
\quad
\subfigure[REFIT House 2]{
\includegraphics[width=3.3cm]{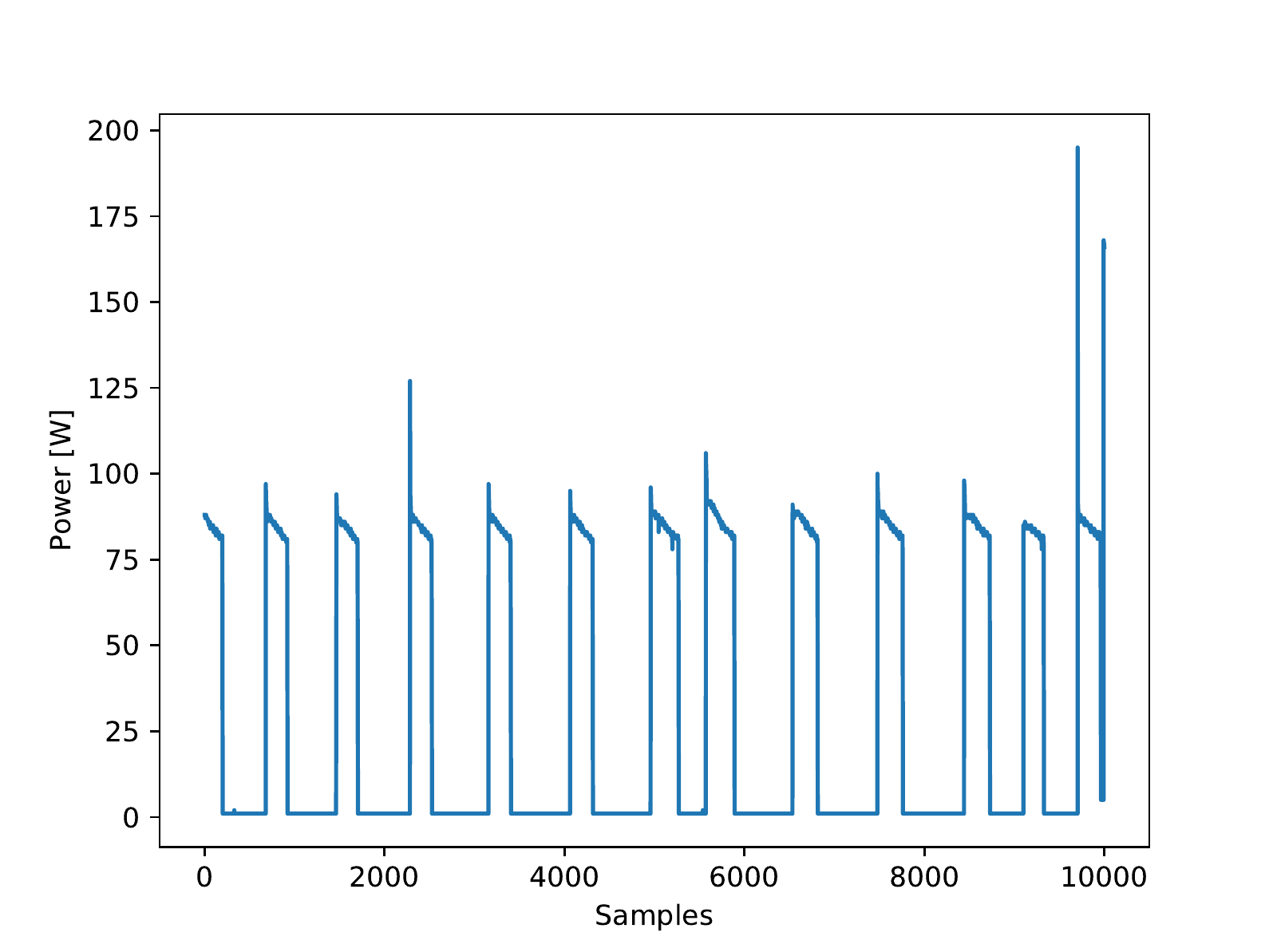}
}
\quad
\subfigure[REFIT House 5]{
\includegraphics[width=3.3cm]{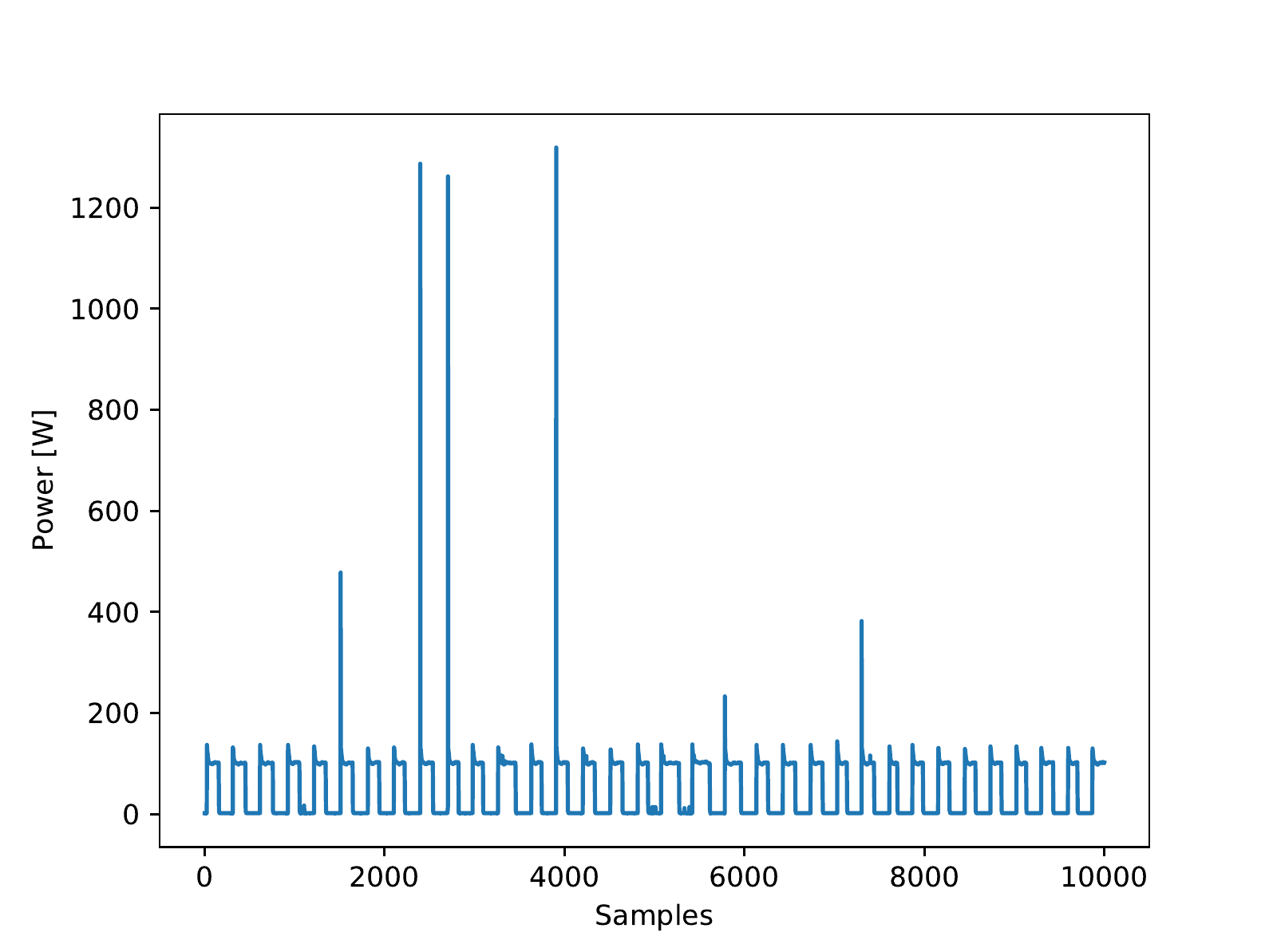}
}
\quad
\subfigure[REFIT House 9]{
\includegraphics[width=3.3cm]{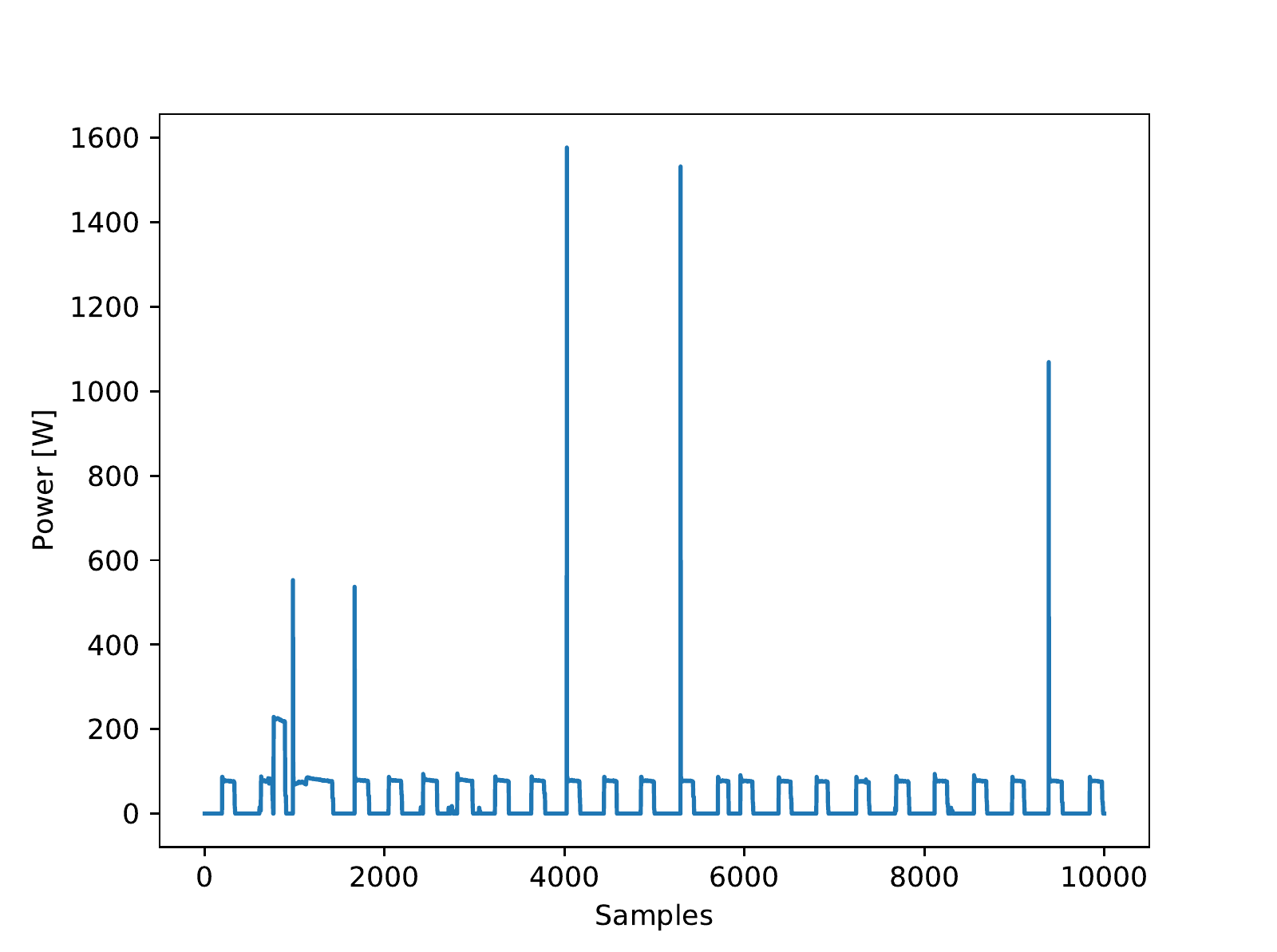}
}
\caption{Example fridge usage distributions for UKDALE, REDD, and REFIT.}
    \label{FigDataExample}
\end{figure}

Based on the evaluation results in the above subsection, although the performance of FedAvg-NILM in the case of 9 smart meter clients achieved satisfying performance on dishwasher and washing machine, its scores on the fridge are worse than both the Local-NILM and the Centralize-NILM models. We further conjecture that the load consumption distribution of the fridge for the clients may be heterogeneous because they are collected geographically, i.e., the REDD dataset is from the U.S., whereas the other two datasets are from the U.K., and the size of the smart meter records from REDD are smaller than the other two datasets. Figure \ref{FigDataExample} shows the load consumption distribution of the fridge for all 9 clients. It can be seen that different clients have different fridge usage patterns, and in particular, the UKDALE and REFIT datasets differ significantly from those in REDD.

We hypothesize that using optimisation algorithms that can accommodate statistical heterogeneity may be useful for improving the performance of FL models. In this subsection, we will explore the relationship between data heterogeneity and the different types of FL utility optimisation models. By comparing the FedProx-NILM to the FedAvg-NILM, we evaluate the ability of the two strategies to learn from heterogeneous data in the DP$^2$-NILM framework. Table \ref{tabFedProx-NILM} lists the average performance scores of the FedAvg-NILM and the FedProx-NILM, and we highlight the improvement in blue and the downgrade in red of the FedProx-NILM corresponding to the FedAvg-NILM. 
\renewcommand\arraystretch{1.2}
\begin{table}[H]
	\centering
		\caption{Average performance scores of FedAvg-NILM and FedProx-NILM schemes for 9 households}
		\begin{center}
		\resizebox{\textwidth}{!}{
\begin{tabular}{ccccccccccccc}
\hline
                      & \multicolumn{4}{c}{\textbf{Fridge}}                                             & \multicolumn{4}{c}{\textbf{Dishwasher}}                                        & \multicolumn{4}{c}{\textbf{Washing Machine}}                                \\ \cline{2-13} 
                      & Accuracy          & F$_1$             & Precision         & Recall              & Accuracy          & F$_1$             & Precision           & Recall            & Accuracy            & F$_1$               & Precision & Recall              \\ \hline
\textbf{FedAvg-NILM}  & 0.65              & 0.63              & 0.50              & 0.85                & 0.97              & 0.75              & 0.92                & 0.64              & 0.98                & 0.71                & 0.83      & 0.62                \\ \hline
\textbf{FedProx-NILM} & 0.85              & 0.81              & 0.82              & 0.81                & 0.98              & 0.80              & 0.78                & 0.82              & 0.97                & 0.54                & 0.83      & 0.40                \\ \hline
\textbf{Evaluation}   & (\textcolor{blue}{$\uparrow$ 20\%}) & (\textcolor{blue}{$\uparrow$ 18\%}) & (\textcolor{blue}{$\uparrow$ 32\%}) & (\textcolor{purple}{$\downarrow$ 4\%}) & (\textcolor{blue}{$\uparrow$ 1\%}) & (\textcolor{blue}{$\uparrow$ 5\%}) & (\textcolor{purple}{$\downarrow$ 19\%}) & (\textcolor{blue}{$\uparrow$ 18\%}) & (\textcolor{purple}{$\downarrow$ 1\%}) & (\textcolor{purple}{$\downarrow$ 17\%}) & (-)       & (\textcolor{purple}{$\downarrow$ 22\%}) \\ \hline
\end{tabular}
			}
			\label{tabFedProx-NILM}
	\end{center}
\end{table}

It can be observed that FedProx-NILM significantly outperforms FedAvg for most scores on fridge and dishwasher, especially on fridge with an improved accuracy by 20\% and an increased precision by 32\%. On the other hand, there are a slightly drop (1\%) on accuracy and a significant drop (22\%) on recall of the washing machine. We infer that, as the signatures of the washing machines are more complex than the other two appliances \cite{d2019transfer}, the proximal term in FedProx-NILM reduces the difference in weight updates for individual models, which may undermine the learning of significant features of washing machines by the client models. To conclude, the above results further confirm our assumptions regarding the utility optimization based on FedProx-NILM for handling heterogeneous smart meter appliances. 

\subsection{Evaluations on privacy-preserving of DP$^2$-NILM}
FedAvg-NILM and FedProx-NILM presented unique advantages for devices with different signatures and datasets with different degrees of consistency. However, studies are suggesting that potential risks still exist in the training communication process even though the transmitted objects are the updated parameters instead of the original data \cite{carlini2019secret}. Therefore, it is necessary to provide stronger privacy guarantees to the FL-based NILM. In this subsection, we evaluate two privacy-preserving schemes of DP$^2$-NILM, i.e., the GDPFL-NILM and the LDPFL-NILM. When clients decide whether to participate in the DP$^2$-NILM paradigm for smart meter data analysis, our framework serves as a reference for quantifying the potential privacy loss based on the privacy budget $\epsilon$. By comparing the benefits of participating in the framework, clients can make an informed decision on whether to join. 

Table \ref{tabPrivacy} compares the GDPFL-NILM and the LDPFL-NILM trained with varied privacy budget $\epsilon$, in which we again use the FedAvg-NILM as the baseline model. Intuitively, the Gaussian random noise will slow the convergence of both the GDPFL-NILM and the LDPFL-NILM models, while providing stronger privacy guarantees for the local clients, leading to trade-off problems between model utility and privacy. It is noticed that even with a stronger privacy guarantee, the GDPFL-NILM still performs reasonable with a marginal reduction in accuracy score from 0.65 to 0.58 with the privacy budget $\epsilon=8$. Interestingly, most of the model performance scores for the dishwasher and washing machine drop dramatically. This is likely because they differ from the fridge in terms of features, as dishwashers and washing machines may offer more insight into individual behaviour because they are more closely related to the routines of smart meter clients.
\renewcommand\arraystretch{1.5}
\begin{table}[H]
	\centering
		\caption{Average performance scores of the GDPFL-NILM and the LDPFL-NILM schemes for 9 households}
		\begin{center}
		\resizebox{\textwidth}{!}{
\begin{tabular}{cccccccccccccccc}
\hline
                                     & \multirow{2}{*}{\textbf{\begin{tabular}[c]{@{}c@{}}Privacy \\ Budget\end{tabular}}} & \multicolumn{4}{c}{\textbf{Fridge}}   & \multicolumn{4}{c}{\textbf{Dishwasher}} & \multicolumn{4}{c}{\textbf{Washing Machine}} & \multirow{2}{*}{\textbf{\begin{tabular}[c]{@{}c@{}}Privacy \\ Guarantee\end{tabular}}} & \multirow{2}{*}{\textbf{\begin{tabular}[c]{@{}c@{}}Trusted \\ Server\end{tabular}}} \\ \cline{3-14}
                                     &                                                                                                   & Accuracy & F$_1$ & Precision & Recall & Accuracy  & F$_1$  & Precision  & Recall & Accuracy   & F$_1$   & Precision   & Recall  &                                                                                        &                                                                                     \\ \hline
\textbf{FedAvg-NILM}                 & $\diagdown$                                                                                        & 0.65     & 0.63  & 0.50      & 0.85   & 0.97      & 0.75   & 0.92       & 0.64   & 0.98       & 0.71    & 0.83        & 0.62    & Basic                                                                                  & Yes                                                                                 \\ \hline
\multirow{3}{*}{\textbf{GDPFL-NILM}} & 4                                                                                                 & 0.54     & 0.53  & 0.37      & 0.95   & 0.90      & 0.14   & 0.16       & 0.72   & 0.95       & 0.68    & 0.40        & 0.92    & \multirow{3}{*}{Moderate}                                                              & \multirow{3}{*}{Yes}                                                                \\
                                     & 8                                                                                                 & 0.63     & 0.61  & 0.49      & 0.84   & 0.97      & 0.69   & 0.93       & 0.56   & 0.98       & 0.68    & 0.79        & 0.60    &                                                                                        &                                                                                     \\
                                     & 12                                                                                                & 0.66     & 0.82  & 0.81      & 0.83   & 0.99      & 0.85   & 0.86       & 0.85   & 0.98       & 0.74    & 0.80        & 0.63    &                                                                                        &                                                                                     \\ \hline
\multirow{3}{*}{\textbf{LDPFL-NILM}} & 4                                                                                                 & 0.58     & 0.40  & 0.40      & 0.38   & 0.93      & 0.11   & 0.21       & 0.39   & 0.94       & 0.10    & 0.11        & 0.34    & \multirow{3}{*}{Strong}                                                                & \multirow{3}{*}{No}                                                                 \\
                                     & 8                                                                                                 & 0.58     & 0.42  & 0.41      & 0.44   & 0.94      & 0.20   & 0.30       & 0.40   & 0.96       & 0.20    & 0.40        & 0.47    &                                                                                        &                                                                                     \\
                                     & 12                                                                                                & 0.65     & 0.42  & 0.36      & 0.50   & 0.94      & 0.13   & 0.26       & 0.48   & 0.96       & 0.43    & 0.40        & 0.50    &                                                                                        &                                                                                     \\ \hline
\end{tabular}
			}
			\label{tabPrivacy}
	\end{center}
\end{table}

\begin{figure}[!htbp]
	\centering
	\includegraphics[width=0.7\textwidth]{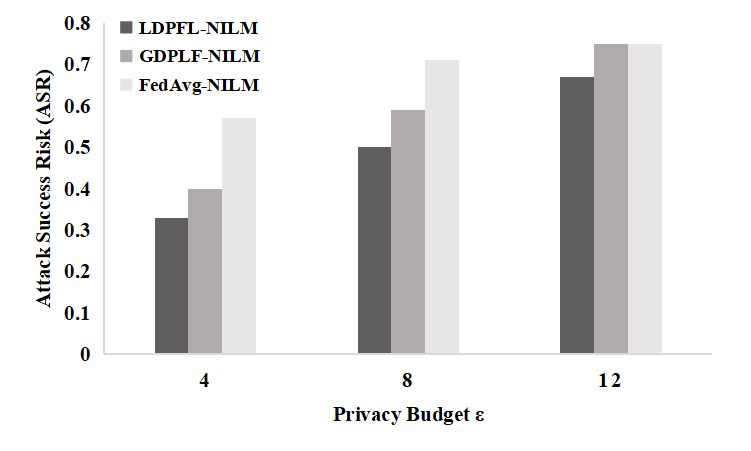}
	\caption{The ASRs of FedAvg-NILM, the GDPFL-NILM, and the LDPFL-NILM in the DP$^2$-NILM framework.}
	\label{DP2NILM_FigASR}
\end{figure}

We then compare the performance of the GDPFL-NILM and the LDPFL-NILM in terms of privacy attacks. To determine whether a client has participated in a training session, we use the attack success risk introduced in Section \ref{EC} as the evaluation criterion. Figure \ref{DP2NILM_FigASR} illustrates ASRs based on various epsilon budgets for FedAvg-NILM, GDPFL-NILM, and LDPFL-NILM, respectively.

We evaluate all the three models with three privacy budget values (i.e., $\epsilon=4$, $\epsilon=8$, and $\epsilon=12$) with a fixed $\delta=10^{-5}$. Figure \ref{DP2NILM_FigASR} shows that LDPFL-NILM with the setting $\epsilon=4$ mitigates the attack success risk better (downgrades the risk to 0.33) with compromises in decreasing model accuracy by 7\% for fridge, 4\% for dishwasher, and 4\% for washing machine. The GDPFL-NILM with $\epsilon=8$ achieved satisfying performance on all the three appliances as well as reduced the attack accuracy to 0.59. 

Not surprisingly, the LDPFL-NILM imposes more noise compared to the GDPFL-NILM, which provides stronger privacy guarantees but less utility due to a higher amount of noise. It is worth noting that with a higher privacy budget $\epsilon=12$, the attack success risk in both the GDPFL-NILM and the LDPFL-NILM are similar to that in FedAvg-NILM whereas the F$_1$, precision, and recall for the LDPFL-NILM are much worse than the FedAvg-NILM and the GDPFL-NILM. Therefore, it may suggest that utilizing the GDPFL-NILM or the FedAvg-NILM may achieve a better trade-off between utility and privacy when there is a higher privacy budget from clients. 

\section{Conclusion}\label{C}
In this paper, we proposed the DP$^2$-NILM framework based on federated learning and differential privacy for NILM, which provides two schemes to the smart meter clients, i.e., the utility optimization scheme and the privacy-preserving scheme. The utility optimization scheme consists of the FedAvg-NILM and the FedProx-NILM focusing on dealing with the data heterogeneity, and the privacy-preserving scheme includes the global differential privacy federated learning NILM and the local differential privacy federated learning NILM to provide privacy guarantees from various levels. We conducted extensive evaluations for the proposed DP$^2$-NILM framework based on real-world smart meter datasets and demonstrated its scalability from multiple perspectives.

The proposed DP$^2$-NILM framework will serve as a key technology to enable electricity utilities and service providers offer various large-scale smart energy services at the local/residential level, thereby enhancing the return on investment from the smart meters. Meanwhile, this framework enables smart meter clients to receive more valuable feedback, which in turn facilitates behaviour change of energy use, and therefore contributes to the decarbonization of the energy system.


This framework has the potential to accommodate a wider range of client types, such as commercial and industrial clients. Nevertheless, the training environment for such client types may be more complex, so it is important to further consider the system heterogeneity to ensure the robustness of the framework. Furthermore, as smart devices enable real-time feedback from smart meter clients, adapting the DP$^2$-NILM framework to online scenarios will deliver more flexible smart meter data analysis and improve the communication efficiency of the FL paradigm.


\bibliography{FDL}

\end{document}